\documentclass[10pt,twocolumn,letterpaper]{article}

\usepackage{cvpr}
\usepackage{times}
\usepackage{epsfig}
\usepackage{graphicx}
\usepackage{amsmath}
\usepackage{amssymb}
\usepackage{multirow}

\usepackage{graphicx}
\usepackage{subcaption}
\usepackage{mwe}
\usepackage{float}

\DeclareMathOperator*{\argmax}{arg\,max}
\DeclareMathOperator{\MSE}{MSE}

\usepackage[dvipsnames]{xcolor}
\usepackage[pagebackref=true,breaklinks=true,letterpaper=true,colorlinks,bookmarks=false]{hyperref}

\cvprfinalcopy % *** Uncomment this line for the final submission

\def\cvprPaperID{} % *** Enter the CVPR Paper ID here

\ifcvprfinal\fi
\begin{document}

%%%%%%%%% TITLE
\title{Training Deep Normalizing Flow Models in Highly Incomplete Data Scenarios with Prior Regularization}

\author{Edgar A. Bernal\\
Rochester Data Science Consortium\\
University of Rochester\\
{\tt\small edgar.bernal@rochester.edu}}

\maketitle
%\thispagestyle{empty}

%%%%%%%%% ABSTRACT
\begin{abstract}
Deep generative frameworks including GANs and normalizing flow models have proven successful at filling in missing values in partially observed data samples by effectively learning --either explicitly or implicitly-- complex, high-dimensional statistical distributions. In tasks where the data available for learning is only partially observed, however, their performance decays monotonically as a function of the data missingness rate. In high missing data rate regimes (e.g., $60\%$ and above), it has been observed that state-of-the-art models tend to break down and produce unrealistic and/or semantically inaccurate data. We propose a novel framework to facilitate the learning of data distributions in high paucity scenarios that is inspired by traditional formulations of solutions to ill-posed problems. The proposed framework naturally stems from posing the process of learning from incomplete data as a joint optimization task of the parameters of the model being learned and the missing data values. The method involves enforcing a prior regularization term that seamlessly integrates with objectives used to train explicit and tractable deep generative frameworks such as deep normalizing flow models. We demonstrate via extensive experimental validation that the proposed framework outperforms competing techniques, particularly as the rate of data paucity approaches unity. 
\end{abstract}
\vspace{-0.5cm}
%%%%%%%%% BODY TEXT
\section{Introduction}

Deep generative models (DGMs) have enjoyed success in tasks involving the estimation of statistical properties of data.  Applications of DGMs involve generation of high-resolution and realistic synthetic data \cite{brock2018large,NIPS2014_5423,miyato2018spectral,oord2016wavenet,NEURIPS2019_5f8e2fa1}, exact \cite{45819,NIPS2018_8224} and approximate \cite{kingma2013autoencoding,pmlr-v32-rezende14} likelihood estimation, clustering \cite{DBLP:journals/corr/abs-1809-04747}, representation learning \cite{pmlr-v80-achlioptas18a, yang2017deep}, and unsupervised anomaly detection \cite{Kiran_2018}. Fundamentally, generative models perform explicit and/or implicit data density estimation \cite{DBLP:journals/corr/Goodfellow17}. Given the complexity of most signals of interest to the learning community (e.g., audio, language, imagery and video), reliably learning the statistical properties of a given population of data samples often requires immense amounts of training data. Recent work has empirically shown that, in order to continue pushing the state-of-the-art in high-fidelity synthetic data generation, scalable models able to ingest ever-growing data sources may be required \cite{brock2018large}.

Some of the data requirements imposed by current deep generative models may limit their applicability in real-life scenarios, where available data may not be plentiful, and additionally, may be noisy, or only partially observable. The nature of real-world data poses challenges to existing models, and mechanisms to overcome those challenges are needed in order to further the penetration of the technology. In this paper, we focus on enabling the learning of DMGs in scenarios of high data missingness rates (e.g., 60\% of entries missing per data sample and above), where the missingness affects \textit{both} the training and the test sets.  We specifically focus on the task of image imputation, which consists in filling in missing or unobserved values without access to fully observed images during training.  Previous work on data imputation leveraging various forms of DMGs has explicitly addressed image imputation \cite{li2018learning,icml2020_3129,Richardson_2020_CVPR,pmlr-v80-yoon18a}.  While the results are reasonable in low- and mid-data missingness regimes, empirical results indicate that, as large fractions of the data become unobserved, either the perceived quality of the recovered data suffers \cite{Richardson_2020_CVPR}, the original semantic content in the image is lost \cite{pmlr-v80-yoon18a} or both \cite{li2018learning}. These undesired consequences are likely caused by the ill-posedness of the problem of attempting to estimate certain statistical properties from partially observed data, an issue which becomes more extreme as the rate of unobserved data approaches totality. Of note, most existing work fails to consider the semantic content preservation aspect of the task altogether, and focuses solely on measuring the performance of the algorithms based on the realism of the recovered data samples \cite{li2018learning,icml2020_3129,pmlr-v80-yoon18a}. 

Inspired by these observations, we propose to constrain the complexity of the solution space where the reconstructed image lies via regularization techniques, a technique initially exploited in traditional ill-posed inverse problem formulations \cite{Tikhonov/Arsenin/77} and more recently adapted to statistical learning scenarios \cite{Vapnik1998}. The proposed regularization term enforces a prior distribution on the gradient map of the reconstructed images \cite{NIPS2009_3707} in the form of a shallow, hand-engineered constraint, and stands in contrast with recent trends which rely on the high expressivity and capacity of deep models to effectively construct data-driven priors, but which break down in scenarios where data scarcity is an issue.  We seamlessly couple the regularizing priors with explicit likelihood estimates of reconstructed samples yielded by normalizing flows in a novel framework we dub \textit{PRFlow}, which stands for \textit{P}rior-\textit{R}egularized Normalizing \textit{Flow}. The contributions of this paper are as follows:

\vspace{-0.2cm}
\begin{itemize}
	\setlength\itemsep{-0.15cm}
	\item a framework combining traditional explicit and tractable deep generative models with shallow, hand-engineered priors in the form of regularization terms to constrain the complexity of the solution space in high data paucity regimes;
	\item a formal derivation of the framework stemming from the formulation of the learning task with incomplete data as a joint optimization task of the network parameters and missing data values; 
%	\item derivation of how to couple the shallow regularizing prior into a deep generative framework by leveraging the explicit likelihood estimation capabilities of normalizing flow models;
	\item a comprehensive testing framework --including a new metric that captures the semantic consistency between the original and the recovered data samples-- which evaluates all aspects of performance that are relevant when learning from partially observed data; and
	\item empirical validation of the effectiveness of the proposed framework on the imputation of three standard image datasets and benchmarking against current state-of-the-art imputation models under the proposed testing framework.
	
\end{itemize}\vspace{-0.3cm}
%-------------------------------------------------------------------------
\section{Related Work}

Deep learning frameworks have proven successful at a wide range of applications such as speech recognition, image and video understanding, and game playing, but are often criticized for their data-hungry nature \cite{DBLP:journals/corr/abs-1801-00631}.  Some scholars go as far as to say that the future of deep learning depends on data efficiency, and have attempted to achieve it in various ways, for example, by leveraging common sense \cite{10.1145/3186549.3186562}, mimicking human reasoning \cite{Georgeeaag2612} or incorporating domain knowledge into the learning process \cite{8621955}.  The ability to learn from incomplete, partially observed and noisy data will be fundamental to advance the adoption of deep learning frameworks in real-life applications.  In recent years, a body of research on deep frameworks that can learn from partially observed data has emerged.  Initial work focused on extensions of generative models such as Variational Auto Encoders (VAEs) \cite{kingma2013autoencoding} and Generative Adversarial Networks (GANs) \cite{NIPS2014_5423}, including Partial VAEs \cite{Ma2018PartialVF}, the Missing Data Importance-Weighted Autoencoder (MIWAE) \cite{pmlr-v97-mattei19a}, the Generative Adversarial Imputation Network (GAIN) \cite{pmlr-v80-yoon18a} and the GAN for Missing Data (MISGAN) \cite{li2018learning}.  More recently, the state-of-the-art benchmark on learning from incomplete data has been pushed by bidirectional generative frameworks which leverage the ability to map back and forth between the data space and the latent space.  Two such examples include the Monte-Carlo Flow model (MCFlow) \cite{Richardson_2020_CVPR} which relies on explicit normalizing flow models \cite{Dinh2014NICENI, 45819, NIPS2018_8224}, and the Partial Bidirectional GAN (PBiGAN) \cite{icml2020_3129} which extends the bidirectional GAN framework \cite{DBLP:conf/iclr/DonahueKD17,dumoulin2017adversarially-iclr}.

While the results achieved by recent work are impressive in their own right, these methods share a common thread: they all break down, in one way or another, as the missingness rate in the data approaches unity.  This phenomenon can be intuitively understood if we think of a generative model as a probability density estimator (either explicit or implicit) \cite{DBLP:journals/corr/Goodfellow17}, which is, at its core, an ill-posed inverse problem \cite{doi:10.1111/1467-9574.00114, rosenblatt1956}.  From this standpoint, the ill-posedness becomes more extreme as the rate of occurrence of unobserved data increases.  Historically, regularization techniques \cite{engl1996regularization,Tikhonov/Arsenin/77} have been widely used to precondition estimators and avoid undesired behaviors of solutions by restricting the feasible space \cite{Evgeniou2000,Vapnik1998}.  While regularization in deep learning is commonplace (e.g., weight decay and weight sharing \cite{10.1162/neco.1992.4.4.473}, dropout \cite{10.5555/2627435.2670313}, batch normalization \cite{pmlr-v37-ioffe15}), it is usually implemented to constrain the plausible space of network parameters and avoid overfitting in discriminative scenarios.  Models that implement regularization on the \textit{output} space tend to be of the generative type. For instance, image priors have been leveraged to address the inherently ill-posed single-image super resolution problem \cite{4587659,8099502,5539933,7410407,zhang2020super}.  The proposed framework can be seen as an attempt to incorporate domain knowledge in learning scenarios in order to guide, facilitate or expedite the learning \cite{10.5555/1756006.1859918, hu-etal-2016-harnessing, NIPS2018_8250}.

\section{Proposed Framework}
%\subsection{Learning from Incomplete Data as an Ill-Posed Problem}
Parallels between ill-posed inverse problems and learning tasks have been established in the literature \cite{NIPS2004_2722,Vapnik1998}.  To informally  illustrate how the degree of ill-posedness of a learning task from partial observation grows with the rate of data missingness, consider the task of image imputation.  Let $b$ denote the bit depth used to encode each pixel value (i.e., pixels can take on values $g$, where $0 \leq g \leq 2^b - 1$) and $N$ the number of pixels of the images in question.  The total number of possible images that can be represented with this scheme is $(2^b)^N$. For the sake of discussion, let us ignore the fact that natural images actually lie on a lower-dimensional manifold within that image space.  Let $0 \leq p \leq 1$ denote the data missingness rate.  This means that when we partially observe an image, we are only exposed to $(1 - p) \cdot N$ of its pixel values.  The task of image imputation involves estimating the remaining $p \cdot N$ pixel values, which means that for every partially observed input image, there are $(2^b)^{pN}$ possible imputed solutions.  It is apparent that the dimensionality of the feasible solution space grows exponentially as the missingness rate grows linearly.  The practical implication of this observation is that, in order to maintain a certain level of reconstruction performance, the number of partially observed data samples needs to grow exponentially as the missingness rate grows linearly. This is an example of an ill-posed problem where the observed data itself is not sufficient to find unique solutions. %In order to quantify the degree of uncertainty involved in the imputation task, multiple imputation schemes have been proposed \cite{little2002statistical}.

When an imputation task is tackled with a learning framework (i.e., a deep generative network), the inductive bias that arises from the choice of network inherently constraints the solution space.  This restriction is not only convenient but also necessary for learning \cite{DBLP:conf/iclr/CohenS17}, as illustrated by recent work which shows that the structure of a network captures natural image statistics prior to any training \cite{Ulyanov_2018_CVPR}.  We will demonstrate empirically that inductive bias alone is not sufficiently effective at restricting the solution space in cases where data is missing at high rates.  Experimental results conclusively show that augmenting the constraining properties of the inductive bias with shallow priors implemented in the form of regularizers is a simple an effective strategy in boosting the performance of deep models in scenarios of high data paucity.

\subsection{Framework Description}\label{sec:prior}

Normalizing flow models are explicit generative models which perform tractable density estimation of the observed data.  The density estimate is constructed by learning a cascade of invertible transformations which perform a mapping between the data space and a latent space. A simple, continuous prior is assumed on the latent variables, for example a spherical Gaussian density. Exact log-likelihood computation is achieved using the change of variable formula \cite{Dinh2014NICENI, 45819, NIPS2018_8224}.  In this work, we introduce a principled framework that leverages the explicit and tractable likelihood capabilities of normalizing flow models to impose structured constraints on the constructed probabilistic models.

Although the proposed framework is generic enough to support a wide range of prior constraints, this study leverages the Hyper-Laplacian prior \cite{NIPS2009_3707}, which has been proven effective at modelling the heavy-tailed nature of the distribution of gradients in natural scenes. This distribution takes on the form $p_p(z) \propto e^{-k |z| ^ \alpha}$ (or equivalently, $\log p_p(z) \propto -k |z| ^ \alpha$), where $0<\alpha\leq1$ determines the heaviness of the tails in the distribution, and $z$ is the gradient map of image $x$, which can be obtained by convolving $x$ with a family of kernels $f_i$. Subscript $p$ is used to denote the nature of the distribution (i.e., to contrast with data-driven priors). We use the notation $z = x * f_i$ to denote the convolution between image $x$ and kernel $f_i$.  When multiple filters are used, it is common to assume independence of the different edge maps so that $\log p_p(x) \propto - \sum_{i=1}^{I} |x * f_i|^\alpha$, where $I$ is the total number of filters.  

In scenarios where training data is only partially observed, training a normalizing flow model can be formulated as a joint optimization task where two sets of parameters are learned concurrently, namely the missing entries in the data and the parameters of the normalizing flow model itself.  Let $x_{rec}$ denote the reconstructed samples and $G_\theta$ the normalizing flow network parameterized by $\theta$.  The objective of the learning task can be written as

\vspace{-0.2cm}
\begin{equation}
%x_{rec} = \argmax\limits_{x} \left\{p(x|x_{obs})\right\}\\
(x_{rec},\theta^*) = \argmax\limits_{x,\theta} \left\{p(x,\theta)\right\}\\
\label{eq:objective2}
\end{equation}
\vspace{-0.3cm}

Note that, as per the above objective, missing data values are treated as parameters to optimize. Throughout the remainder of the paper, we will refer to these values interchangeably as data parameters or missing data values. 

Estimating the joint density from Eq.~\ref{eq:objective2} is difficult.  One way to circumvent this obstacle is to alternately optimize over the conditional distributions of each of the parameters interest, in a manner similar to the way sampling-based optimization frameworks such as Gibbs Sampling and MCMC \cite{TakahashiMCMC} operate.  Following this principle, the joint optimization task can be broken down into two conditional optimization tasks of likelihood functions.  On the one hand, learning the parameters $\theta$ of normalizing flow network $G_{\theta}$ can be achieved in the traditional manner, that is, by maximizing the log-likelihood of the observed data:

\vspace{-0.2cm}
\begin{equation}
%\theta^* = \argmax\limits_{\theta} \left\{p(\theta|x_{rec},x_{obs})\right\}\\
\theta^* = \argmax\limits_{\theta} \left\{p(\theta|x_{rec})\right\}\\
\label{eq:flow}
\end{equation}
\vspace{-0.3cm}

A set of parameters $\theta$ defines an invertible network $G_{\theta}$ that maps images to a tractable latent space and vice-versa.  Specifically, in order to perform log-likelihood estimation, a data sample $x_i$ is mapped to its latent representation $y_i$ by passing it through $G_\theta$, namely $y_i = G_\theta(x_i)$. Since the likelihood for $y_i$ is known (e.g., from a normality assumption), $p(x_i)$ (i.e., the likelihood of $x_i$) can be computed exactly via the variable change rule.  The ability to estimate the likelihood of a data sample enables the resolution of the second conditional optimization task, which aims at finding the optimal entries for the missing values in the partially observed data by maximizing the likelihood of the reconstructed sample conditioned on the current model parameters:

\vspace{-0.2cm}
\begin{equation}
%x_{rec} = \argmax\limits_{x} \left\{p(x|x_{obs})\right\}\\
x_{rec} = \argmax\limits_{x} \left\{p(x|\theta^*)\right\}\\
\label{eq:objective}
\end{equation}
\vspace{-0.3cm}

\noindent where the search space is constrained to images $x$ whose entries match the observed entries of $x_{obs}$. Solving the optimization task from Eq.~\ref{eq:objective} effectively fills in unobserved data values, that is, performs data imputation. Training the overall imputation model involves alternately solving Eqs.~\ref{eq:flow} and \ref{eq:objective}, which yields a sequence of parameter pairs $(x_{rec}^{(n)}, \theta^{*(n)})$.  Convergence is achieved when little change is observed in the updated parameters.  The description of the framework around Eqs.~\ref{eq:flow} and \ref{eq:objective} follows closely the formulation in \cite{Richardson_2020_CVPR}, although in that work, the training of the model was not framed as a joint optimization task.

As stated, solving Eq.~\ref{eq:flow} involves training a traditional normalizing flow model with the current estimate of the data parameters, i.e., the current values of the imputed data.  In contrast, the optimization task in Eq.~\ref{eq:objective} is a highly ill-posed problem when the data missing rate in $x_{obs}$ is high. PRFlow leverages the key insight that regularization of the task with prior knowledge on the solution space leads to improved, more stable solutions to Eq.~\ref{eq:objective}.  In order to incorporate this prior knowledge, first observe that, as per the Bayes rule:

\vspace{-0.2cm}
\begin{equation}
p(x|\theta^*) \propto p(\theta^*|x)p_p(x)
\label{eq:bayes}
\end{equation}

\noindent where $p_p(x)$ is the prior introduced at the beginning of Sec.~\ref{sec:prior}, and it has been assumed that model parameters $\theta^*$ are fixed.  This is the case since at this stage in the training alternation, the optimization is over the missing data entries with the goal of performing data imputation.  Combining Eqs.~\ref{eq:objective} and \ref{eq:bayes} and applying $\log$ yields

\vspace{-0.5cm}
\begin{equation}
%x_{rec} = \argmax\limits_{x} \left\{p(x|x_{obs})\right\}\\
x_{rec} = \argmax\limits_{x} \left\{\log p(\theta^*|x) + \lambda \log p_p(x)\right\}\\
\label{eq:objective1}
\end{equation}
\vspace{-0.5cm}

\noindent where $\lambda$ is a parameter that controls the desired degree of regularization.  In summary, training PRFlow involves alternately optimizing the objectives in Eqs.~\ref{eq:flow} and \ref{eq:objective1}.  It is worthwhile noting that the objective from Eq.~\ref{eq:flow} and the first term in the objective from 
Eq.~\ref{eq:objective1} involve optimizing the same likelihood function relative to two different sets of parameters, namely the model parameters and the missing data values, respectively.

\subsection{Framework Implementation}
\begin{figure*}
\begin{center}
\includegraphics[width=0.8\linewidth]{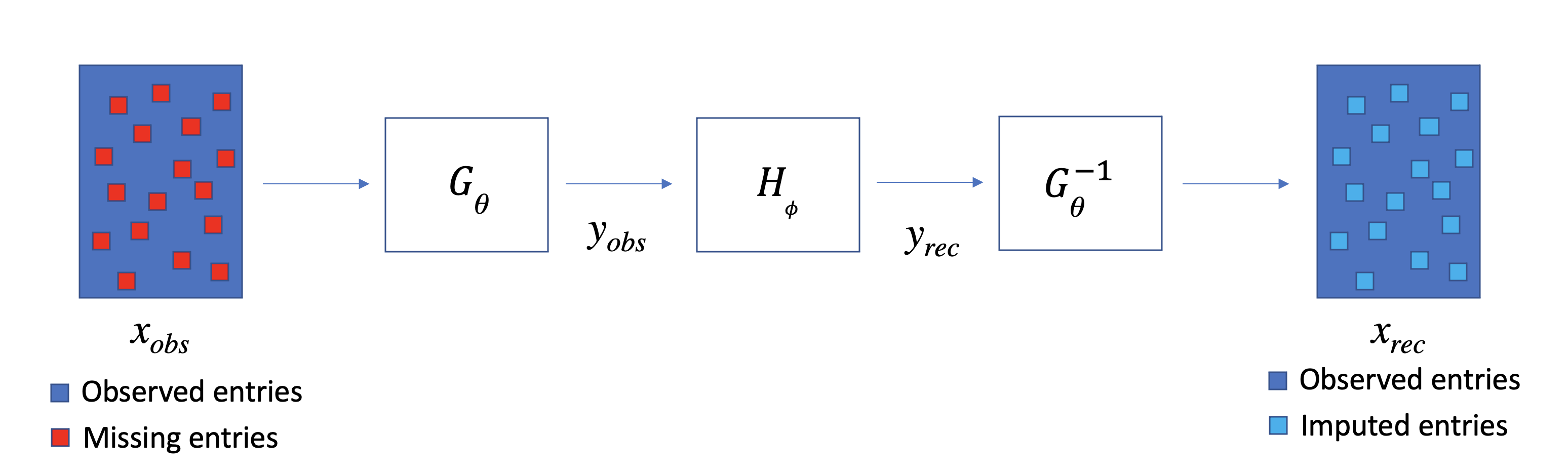}
\end{center}
\vspace{-0.3cm}
   \caption{High-level view of the imputation process.}
\label{fig:flow}
\end{figure*}

PRFlow is largely based on the architecture introduced in \cite{Richardson_2020_CVPR}, which includes a normalizing flow network $G$ that enables likelihood estimation, and a network $H$ performing a non-linear mapping in the latent flow space and fills in missing values in the partially observed data samples. As in \cite{Richardson_2020_CVPR}, network $G$ is an instantiation of RealNVP \cite{45819}. The mapping to the latent space via $G$ is performed because likelihood computation is tractable in that space, and the imputation task is being formulated as the solution of a maximum likelihood conditional objective (as per Eq.~\ref{eq:objective}). At a high level, the imputation process comprises receiving a partially observed sample $x_{obs}$, computing its latent representation $y_{obs} = G_{\theta}(x_{obs})$, mapping this latent representation to $y_{rec} = H_{\phi}(y_{obs})$ with maximum likelihood, and recovering the corresponding maximum likelihood data sample $x_{rec} = G_{\theta}^{-1}(y_{rec})$ which matches the observed entries of $x_{obs}$.  This process is illustrated in Fig.~\ref{fig:flow}. 

As described in Sec.~\ref{sec:prior}, learning this framework involves optimizing two different objectives:~training network $G_{\theta}$ and $H_{\phi}$ involves optimizing the objectives from Eqs.~\ref{eq:flow} and \ref{eq:objective1} respectively, with the optimization being carried out in an alternating way until convergence is achieved. The objectives used to learn these networks, as described below, are denoted $\mathcal{J}(\theta)$ and $\mathcal{J}(\phi)$. In the context of the proposed framework, the data parameters are not optimized directly; instead, network $H_{\phi}$ is learned according to $\mathcal{J}(\phi)$, a proxy objective to that in Eq.~\ref{eq:objective1}. We now describe how the two networks are learned.

Learning the optimal parameters $\theta^*$ of normalizing flow network $G_{\theta}$ is achieved by maximizing the log-likelihood of the training data, or equivalently, minimizing the cost function:

\vspace{-0.5cm}
\begin{equation}
%x_{rec} = \argmax\limits_{x} \left\{p(x|x_{obs})\right\}\\
\mathcal{J}(\theta) = -\sum_i{\log p_\theta(x_{i}^{(n)})}
\label{eq:loss1}
\end{equation}
\vspace{-0.5cm}

\noindent where the sum is computed across training data samples, and the superscript $(n)$ denotes samples which have been imputed with the most recent (i.e., the $n$-th) imputation model. Throughout this optimization stage, the training data remains unchanged. At initialization, where no imputation model is available, shallow imputation techniques (e.g., nearest neighbor or bilinear interpolation) are used. Minimizing this loss corresponds to solving the optimization task from Eqs.~\ref{eq:flow}.

Learning the optimal parameters $\phi^*$ of the imputation model, which operates in the latent space of the normalizing flow network, is achieved by minimizing a three-term loss. Updating parameters $\phi$ results in an updated imputer network $H_\phi$, which is used to obtain an updated training set $x^{(n)}$. Throughout this stage, normalizing flow network $G_\theta$ remains fixed. The first element of the loss involves maximizing the likelihood of the reconstructed samples as per the likelihood estimate provided by the normalizing flow model, or equivalently, minimizing the cost function:

\vspace{-0.5cm}
\begin{equation}
%x_{rec} = \argmax\limits_{x} \left\{p(x|x_{obs})\right\}\\
\begin{split}
\mathcal{J}_1(\phi) &= -\sum_i{\log p_\theta(x_i^{(n)})} \\
& = -\sum_i{\log p_\theta \left[G^{-1}_\theta \circ H_\phi \circ G_\theta(x_i^{(n-1)})\right]}
\end{split}
\label{eq:loss2.1}
\end{equation}
\vspace{-0.5cm}

\noindent where the $\circ$ operator denotes functional composition and the expression for $x_i^{(n)}$ has been expanded to emphasize its dependence on the parameters being optimized, namely $\phi$. Minimizing this loss is equivalent to optimizing the first term of the objective from Eq.~\ref{eq:objective1}.  As stated before (see last paragraph in Sec.~\ref{sec:prior}), this loss is equivalent to the loss from Eq.~\ref{eq:loss1}; the difference lies in the set of parameters that are being modified to achieve the objective. This term encourages the imputer to output recovered samples that are more likely to occur.

The second element involves minimizing the discrepancy between the recovered data and the known entries of the observed data:

\vspace{-0.5cm}
\begin{equation}
\begin{split}
%x_{rec} = \argmax\limits_{x} \left\{p(x|x_{obs})\right\}\\
\mathcal{J}_2(\phi) = \sum_i{\MSE(x_{i,obs},G^{-1}_\theta \circ H_\phi \circ G_\theta(x_i^{(n-1)}))}
\end{split}
\label{eq:loss2.2}
\end{equation}
\vspace{-0.5cm}

\noindent where the MSE is computed across the known entries of the observed data only. Note that these entries remain unchanged throughout both stages of the optimization process, thus no superscript is needed. This term encourages the imputer to output recovered samples that match the known entries at the observed positions.

The last term penalizes reconstructions that deviate from the expected behavior as dictated by the regularizing prior:

\vspace{-0.5cm}
\begin{equation}
\begin{split}
%x_{rec} = \argmax\limits_{x} \left\{p(x|x_{obs})\right\}\\
\mathcal{J}_3(\phi) &= -\sum_i{\log p_p(x_i^{(n)})} = -\sum_i{\sum_j{|x_i^{(n)} * f_j|^\alpha}} \\
&= -\sum_i{\sum_j{|G^{-1}_\theta \circ H_\phi \circ G_\theta(x_i^{(n-1)})* f_j|^\alpha}}
\end{split}
\label{eq:loss2.3}
\end{equation}
\vspace{-0.5cm}

\noindent where the summations indexed by $i$ and $j$ are performed across data samples and gradient kernels, respectively, and we have incorporated the expression for the prior introduced in Sec.~\ref{sec:prior}. Minimizing this loss is equivalent to optimizing the second term in the objective from Eq.~\ref{eq:objective1}.  In our implementation, and for the sake of computational efficiency and simplicity, we use two first-order derivative filters, namely $[1, 1]$ and $[1, 1]^\intercal$. Note that higher-order or learnable filters can be used instead, which would likely result in improved performance.

In summary, training PRFlow involves joint optimization of objectives $\{\mathcal{J}(\theta),\mathcal{J}_1(\phi),\mathcal{J}_2(\phi),\mathcal{J}_3(\phi)\}$ across $\theta$ and $\phi$, where $\theta$ denotes the parameters of the normalizing flow network and $\phi$ denotes the imputer network parameters, i.e., the parameters that ultimately determine how the missing data values are filled in.
% \begin{figure}[t]
% \begin{center}
%   \includegraphics[width=1.0\linewidth]{Flow.png}
% \end{center}
%   \caption{Example of caption.  It is set in Roman so that mathematics
%   (always set in Roman: $B \sin A = A \sin B$) may be included without an
%   ugly clash.}
% \label{fig:flow}
% \end{figure}

%\subsection{Image Quality Trade-Offs}

\section{Experimental Results}

\textbf{Datasets and Procedure.} The efficacy of PRFlow was evaluated on three different standard image datasets, MNIST \cite{lecun-mnisthandwrittendigit-2010}, CIFAR-10 \cite{krizhevsky2009learning} and CelebA \cite{liu2015faceattributes}.  Four different rates of data missingness were tested, from 60\% to 90\% in steps of 10\%. The training procedure follows the principles of recent work proposing models that support and rely purely on partially observed data during the learning phase \cite{li2018learning,icml2020_3129,Richardson_2020_CVPR,pmlr-v80-yoon18a} by training with the dataset resulting from randomly dropping the corresponding percentage of pixels from the images in the standard training set from the respective dataset of interest according to a Bernoulli distribution.  In MNIST, the training set comprises 60,000 $28\times28$-pixel grayscale images, whereas in CIFAR it includes 50,000 $32\times32$-pixel RGB images. Since no standard partition exists for CelebA, we use the first 100,000 images for training and the remaining for testing. We pre-process CelebA images by performing $108\times108$ pixel center cropping and resizing to $32\times32$ pixels. For testing, we adhere to the experimental principle drawn out in \cite{Richardson_2020_CVPR}, where performance is measured on the standard test set of the relevant dataset after having randomly dropped the appropriate fraction of pixel values.  

\textbf{Metrics.} We measure the performance of the algorithms relative to three different metrics, which we believe capture all relevant attributes of data recovered by an algorithm attempting to reconstruct partially observed data: (i) root mean squared error (RMSE), which measures differences between the reconstructed image and the ground truth at the pixel level; (ii) the Fréchet Inception Distance (FID), first proposed to measure the quality of data produced by generative models \cite{heusel2017gans} and which captures population-level similarities; and (iii) the ratio of the classification accuracy of a classifier pre-trained on fully observed training data on the reconstructed data to the accuracy of the same classifier on the fully observed test set.  This metric, which we denote the Semantic Consistency Criterion (SCC), aims at measuring the amount of semantic information preserved by the missing data recovery process. Formally, let $acc_{imp}$ be the performance of the benchmark classifier on an imputed test set and $acc_0$ the performance of the same classifier on the original test set. Then $\text{SCC} = \min\{1,acc_{imp}/acc_0\}$, where the clipping is introduced to handle the unlikely case when $acc_{imp} > acc_0$. Normalization by the baseline classifier performance is done to minimize the impact of the choice of classifier. This overarching experimental framework contrasts with most previous work on generative modelling of incomplete data (\cite{Richardson_2020_CVPR} excepted), which doesn't consider the preservation of semantic content as a metric of performance, and tends to make more emphasis on the realism of the recovered samples than on the pixel-level accuracy \cite{li2018learning,icml2020_3129}. In this work, we consider all three metrics to be equally important, and posit that one of the most salient strengths of the proposed method is that it minimizes the impact of the trade-off between the metrics relative to competing methods. Of note, RMSE is measured between the recovered values and the ground truth values at the unobserved pixel locations in the test set.  This means that not only the pixels but also the full images used to measure the performance of the method are completely unseen by the framework during training, unlike approaches which measure performance on unobserved values within the training set \cite{li2018learning,icml2020_3129}.  Similarly, FID is measured between the recovered test set and the ground truth test set, and SCC performance is measured on the recovered test set imagery.  

% \begin{figure}
%     \centering
%     \begin{subfigure}[b]{0.225\textwidth}
%         \centering
%         \includegraphics[width=\textwidth]{latex/ConfMat_MisGAN.9.png}
%             \vspace{-0.4cm}
%         \caption[]%
%         {{\small MisGAN (SCC=0.35)}}    
%     \end{subfigure}
%     \hfill
%     \begin{subfigure}[b]{0.225\textwidth}  
%         \centering 
%         \includegraphics[width=\textwidth]{latex/ConfMat_PBiGAN.9.png}
%             \vspace{-0.4cm}
%         \caption[]%
%         {{\small PBiGAN (SCC=0.76)}}    
%     \end{subfigure}
%     \vskip\baselineskip
%     \vspace{-0.3cm}
%     \begin{subfigure}[b]{0.225\textwidth}   
%         \centering 
%         \includegraphics[width=\textwidth]{latex/ConfMat_MCFlow.9.png}
%         \caption[]%
%         {{\small MCFlow (SCC=0.75)}}    
%     \end{subfigure}
%     \hfill
%     \begin{subfigure}[b]{0.225\textwidth}   
%         \centering 
%         \includegraphics[width=\textwidth]{latex/ConfMat_PRFlow.9.png}
%         \caption[]%
%         {{\small PRFlow (SCC=0.75)}}    
%     \end{subfigure}
%     \caption{Confusion matrices for MNIST classification by LeNet on different test sets after imputation with competing methods for 90\% missing data rate, and corresponding SCC values.} 
%         \vspace{-0.2cm}
%     \label{fig:SCC}
% \end{figure}

\textbf{Competing Methods.} We benchmark the performance of PRFlow against three methods, namely MisGAN \cite{li2018learning}, PBiGAN \cite{icml2020_3129} and MCFlow \cite{Richardson_2020_CVPR}, which together comprise the state-of-the-art landscape in image imputation tasks across the considered metrics. We used the publicly available code for all three competing methods from their official repositories; we used the code as published for MNIST and made extensions to the code to enable support of CIFAR (no CIFAR versions were publicly available). We use LeNet \cite{Lecun98gradient-basedlearning}, ResNet18 \cite{He2016DeepRL} to compute both SCC and FID on MNIST and CIFAR, respectively. Since CelebA has no classes, we use FaceNet \cite{conf/cvpr/SchroffKP15} to compute FID only.

\textbf{Experimental Setup.} Throughout the experiments, we use $\alpha = 1/3$, a learning rate of $1\times10^{-4}$, and a batch size of $64$. We train until little change is observed in the loss from Eq.~\ref{eq:loss2.2}, as opposed to competing methods which prescribe a set number of epochs to train. $G_\theta$ is a RealNVP \cite{45819} network with six affine coupling layers. We implement $H_{\phi}$ as a 3-hidden layer, fully connected network with $784$ and $1024$ neurons per layer for MNIST and CIFAR/CelebA, respectively, with input and output layers having the same number of neurons as the dimensionality of the images (i.e., $28\times28=784$ for MNIST, and $32\times32\times3=3072$ for CIFAR and CelebA). Although performance is somewhat robust to the choice of $\lambda$, we noticed it did affect convergence speed: too large a value would lead to oscillations and too small a value would lead to slow convergence. As a rule of thumb, we found that a value of $\lambda$ that approximately equalizes the value of $\mathcal{J}_1(\phi)$ (Eq.~\ref{eq:loss2.1}) and the value of $\lambda \mathcal{J}_3(\phi)$ (Eq.~\ref{eq:loss2.3}) worked well. %A Pytorch implementation of the algorithm will be made available online.

\textbf{Results.} Table \ref{tab:RMSE} includes the RMSE results for all competing methods across both datasets and considered data missingness rates. MCFlow and PRFlow perform similarly, while PBiGAN performs the worst, with the gaps in performance being significantly larger for CIFAR. These results are reasonable since neither MisGAN nor PBiGAN enforce an MSE loss explicitly. Table \ref{tab:FID} includes the FID results laid out in a similar fashion.  In this case, PRFlow again outperforms all competing methods, trailed closely by PBiGAN on MNIST, with performance being more even across the field on CIFAR and CelebA. These results highlight the efficacy of the regularizing prior at shaping the statistical behavior of the recovered imagery. Lastly, Table \ref{tab:SCC} includes SCC results. In the MNIST case, MCFlow, PBiGAN and PRFlow perform similarly, with MisGAN trailing by a somewhat significant margin, and with the margin increasing as the missing rate increases. In the CIFAR case, PRFlow outperforms the competition more handily. %The results from the top half of Table \ref{tab:SCC} are visualized in the form of confusion matrices in Fig.~\ref{fig:SCC}, which correspond to classification performance for different versions of the standard MNIST test set by LeNet. The different test sets are obtained by imputing the standard MNIST test set observed at a 90\% missing rate with MisGAN, PBiGAN, MCFlow and PRFlow (Figs.~\ref{fig:SCC}(a), (b), (c) and (d) respectively). It can be seen that the performance of the three top competitors is closely matched, and that typical classification errors are to be expected (e.g., `4' and `7' being mistaken for '9' and vice-versa). 

\vspace{-0.1cm}
\begin{table}[ht]\caption{RMSE between recovered data and ground truth test set, unobserved pixels only (lower is better)} \label{tab:RMSE} 
\vspace{-0.3cm}
\small
\centering
\setlength\doublerulesep{0.5pt}
\begin{tabular}{|c|c|c|c|c|c|}
\cline{3-6}
\multicolumn{2}{c}{} & \multicolumn{4}{|c|}{\textbf{Missing Rate}} \\
\hline
     \textbf{Dataset}       &            \textbf{Method}                    & 0.6      & 0.7      & 0.8      & 0.9      \\ \hline
\multirow{4}{*}{MNIST}                  & MisGAN  & 0.1329 & 0.1561  & 0.1958 & 0.2484         \\ 
                                                  & PBiGAN  & 0.3155 & 0.3121 &0.3045 &0.2844 \\ 
                                                  & MCFlow      &    0.1126     &  0.1300       &  0.1581      & 0.2080 \\  
                                                  & PRFlow  &   \textbf{0.1093}      &   \textbf{0.1243}      &   \textbf{0.1490}      &  \textbf{0.2059}       \\\hline
\multirow{4}{*}{CIFAR}        & MisGAN  &  0.2568  & 0.2814  & 0.3081 & 0.3461         \\ 
                                                  & PBiGAN &0.3380 &0.3443 &0.3623 & 0.4448\\             
& MCFlow      &  0.0921        &  0.1059       &  0.1187       & 0.1460 \\  
                                                  & PRFlow  & \textbf{0.0802}        &  \textbf{0.0919}       & \textbf{0.1102}  &    \textbf{0.1299}   \\  \hline
\multirow{4}{*}{CelebA}        & MisGAN  &  0.2232	& 0.2273 &	0.2404 &	0.2777         \\ 
                                                  & PBiGAN & 0.2894 &	0.3356	& 0.3733	& 0.4230 \\             
& MCFlow      &  0.0793        &  0.0828       &  0.0927       & 0.1189 \\  
                                                  & PRFlow  & \textbf{0.0738}        &  \textbf{0.0813}       & \textbf{0.0924}  &    \textbf{0.1135}   \\  
\hline

\end{tabular}
\end{table}

\begin{table}[ht]\caption{FID between recovered data and ground truth test sets  (lower is better)} \label{tab:FID} 
\vspace{-0.7cm}
\small
\centering
\setlength\doublerulesep{0.5pt}
\begin{tabular}{|c|c|c|c|c|c|}
\cline{3-6}

\multicolumn{2}{c}{} & \multicolumn{4}{|c|}{\textbf{Missing Rate}} \\

\hline
   \textbf{Dataset}         &            \textbf{Method}                   & 0.6      & 0.7      & 0.8      & 0.9      \\ \hline
\multirow{4}{*}{MNIST}            & MisGAN  & 0.8300 & 1.5373 & 3.0956 & 7.9071         \\ 
                                                  & PBiGAN  & 0.1356 & 0.3082 & 0.9927 & 4.2000 \\
& MCFlow      &    0.7840     &  1.3382       &  3.0663       & 8.5047 \\  
                                                  & PRFlow  &    \textbf{0.0959}     &   \textbf{0.2888}      &   \textbf{0.8795}      &  \textbf{3.8759}      \\  
                                                  \hline
\multirow{4}{*}{CIFAR}            & MisGAN  & 0.7299 & 0.8464  &0.9136&  0.9477       \\ 
                                           & PBiGAN  & 0.8743 & 0.9794 & 1.1229 & 1.1308\\ 
                & MCFlow      & 0.4145        &   0.6564      & 0.8777        &  1.0808 \\ &  PRFlow  &    \textbf{0.2928}     &    \textbf{0.5111}     &   \textbf{0.6825}      &   \textbf{0.8437}    \\ \hline
\multirow{4}{*}{CelebA}            & MisGAN  & 0.3085 &	0.3486 &	0.4024 &	0.5693       \\ 
                                           & PBiGAN  & 0.7547 &	0.7861 &	0.8931 &	0.9415\\ 
                & MCFlow      & 0.1225        &   0.1672     & 0.3333        & 0.7587  \\ &  PRFlow  &    \textbf{0.0887}     &    \textbf{0.1481}     &   \textbf{0.2359}      &   \textbf{0.5213}    \\ \hline
\end{tabular}
\end{table}

\begin{table}[ht]\caption{SCC of recovered test set (higher is better)} \label{tab:SCC} 
\vspace{-0.3cm}
\small
\centering
\setlength\doublerulesep{0.5pt}
\begin{tabular}{|c|c|c|c|c|c|}
\cline{3-6}

\multicolumn{2}{c}{} & \multicolumn{4}{|c|}{\textbf{Missing Rate}} \\

\hline
   \textbf{Dataset}         &            \textbf{Method}                   & 0.6      & 0.7      & 0.8      & 0.9      \\ \hline
\multirow{4}{*}{MNIST}                                                             & MisGAN  &0.9423	&0.8763	&0.6964&	0.3489           \\
& PBiGAN & 0.9807 & 0.9619 & 0.9183 & \textbf{0.7602} \\ 
& MCFlow      &  \textbf{0.9872}       &  \textbf{0.9705}       &  \textbf{0.9279}       &  0.7487 \\  
                                                  & PRFlow  &   0.9842      & 0.9693   &   0.9276     &   0.7471    \\  
\hline
\multirow{4}{*}{CIFAR}                 & MisGAN  & 0.4588  & 0.3828  & 0.3364  &   0.2737       \\ 
                                                  & PBiGAN & 0.3717 & 0.3020 & 0.2396 &  0.1757\\
                                                  & MCFlow      &     0.6606    &  0.5194       &   0.3893      &  0.3218 \\  
                                                  & PRFlow  &    \textbf{0.7225} & \textbf{0.5939}        &   \textbf{0.4719}  &  \textbf{0.3559}     \\  
                                                   \hline
\end{tabular}
\end{table}

\begin{figure}
\begin{center}
\includegraphics[width=0.9\linewidth]{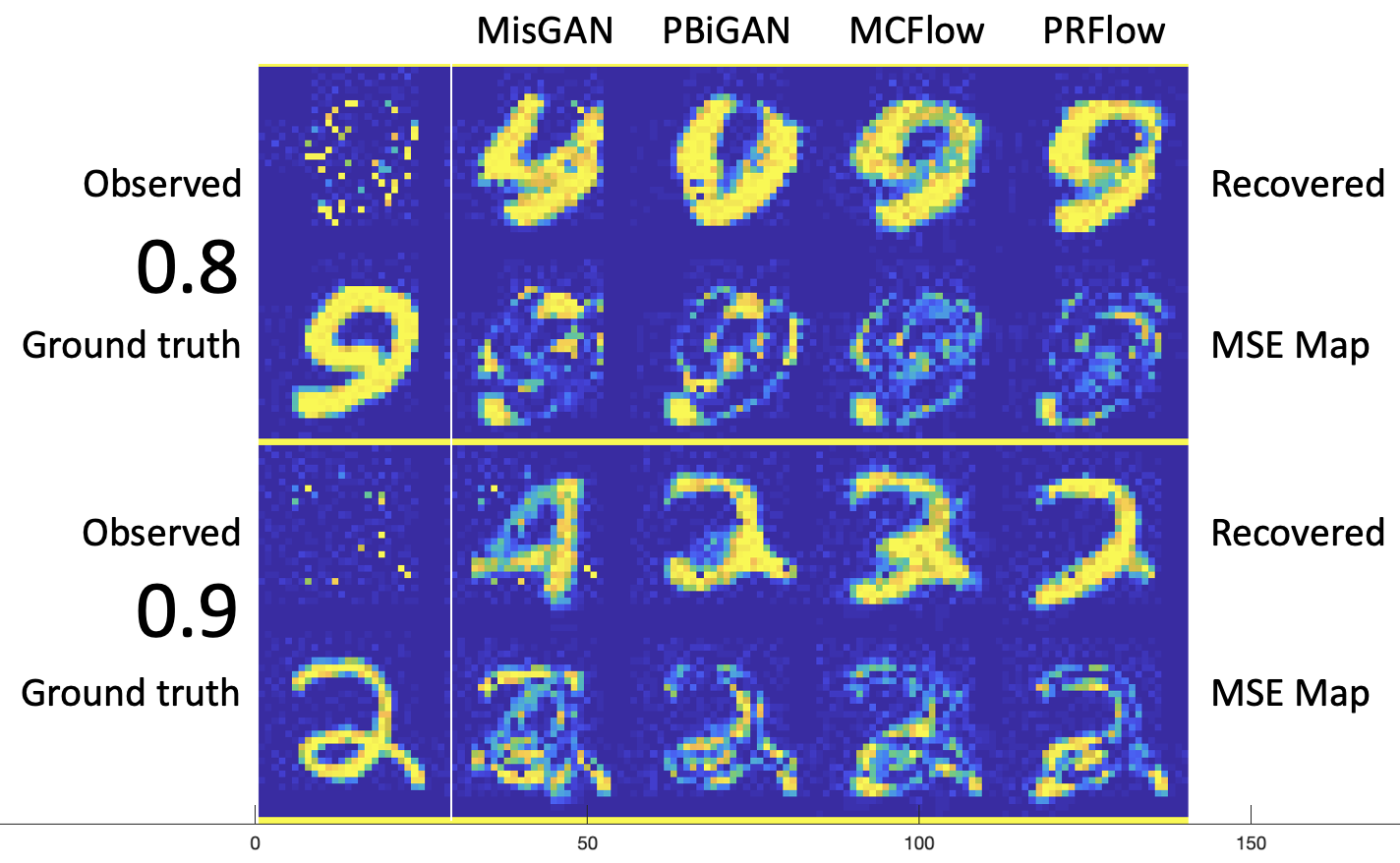}
\end{center}
\vspace{-0.5cm}
   \caption{Sample results on MNIST for 80 and 90\% missing rates (top to bottom image groups).}
\label{fig:mnistresults0}
\end{figure}

\begin{figure}
\begin{center}
\includegraphics[width=0.9\linewidth]{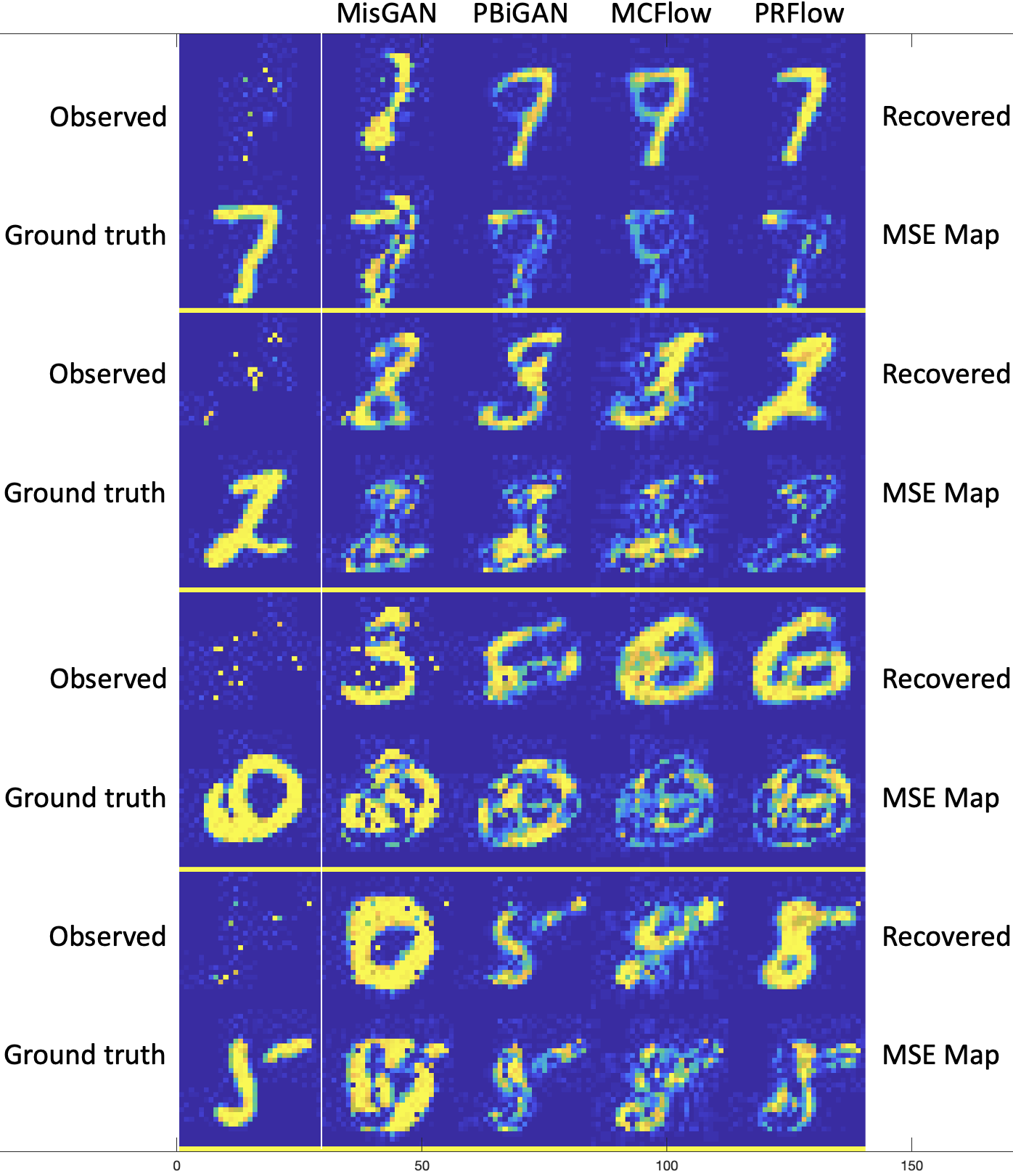}
\end{center}
\vspace{-0.5cm}
   \caption{Sample results on MNIST for 90\% missing data.}
\label{fig:mnistresults1}
\end{figure}

\begin{figure*}
    \centering
    % \begin{subfigure}[b]{0.78\textwidth}
    %     \centering
    %     \includegraphics[width=\textwidth]{latex/impaperidx70_cifar.png}
    % \end{subfigure}
    
    \begin{subfigure}[b]{0.78\textwidth}  
        \centering 
        \includegraphics[width=\textwidth]{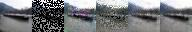}
    \end{subfigure}

    \begin{subfigure}[b]{0.78\textwidth}   
        \centering 
        \includegraphics[width=\textwidth]{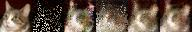}  
    \end{subfigure}
    
    \begin{subfigure}[b]{0.78\textwidth}   
        \centering 
        \includegraphics[width=\textwidth]{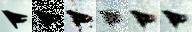}
    \end{subfigure}
     
    % \begin{subfigure}[b]{0.8\textwidth}   
    %     \centering 
    %     \includegraphics[width=\textwidth]{latex/impaperidx495_cifar.png}
    % \end{subfigure}
    
    % \begin{subfigure}[b]{0.78\textwidth}   
    %     \centering 
    %     \includegraphics[width=\textwidth]{latex/impaperidx2385_cifar.png}
    % \end{subfigure}
    \vspace{-0.1cm}
    \caption{Sample results on CIFAR-10. From left to right: ground truth, observed, and MisGAN, PBiGAN, MCFlow and PRFlow reconstructions.} 
\label{fig:cifarresults}
\end{figure*}

\begin{figure*}
    \centering
    % \begin{subfigure}[b]{0.78\textwidth}
    %     \centering
    %     \includegraphics[width=\textwidth]{latex/impaperidx70_cifar.png}
    % \end{subfigure}
    
    \begin{subfigure}[b]{0.78\textwidth}  
        \centering 
        \includegraphics[width=\textwidth]{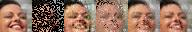}
    \end{subfigure}

    \begin{subfigure}[b]{0.78\textwidth}   
        \centering 
        \includegraphics[width=\textwidth]{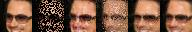}  
    \end{subfigure}
    
    \begin{subfigure}[b]{0.78\textwidth}   
        \centering 
        \includegraphics[width=\textwidth]{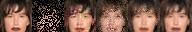}
    \end{subfigure}
    
    \begin{subfigure}[b]{0.78\textwidth}   
        \centering 
        \includegraphics[width=\textwidth]{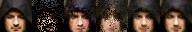}
    \end{subfigure}
    % \begin{subfigure}[b]{0.8\textwidth}   
    %     \centering 
    %     \includegraphics[width=\textwidth]{latex/impaperidx495_cifar.png}
    % \end{subfigure}
    
    % \begin{subfigure}[b]{0.78\textwidth}   
    %     \centering 
    %     \includegraphics[width=\textwidth]{latex/impaperidx2385_cifar.png}
    % \end{subfigure}
    \vspace{-0.1cm}
    \caption{Sample results on CelebA. From left to right: ground truth, observed, and MisGAN, PBiGAN, MCFlow and PRFlow reconstructions.} 
\label{fig:celebaresults}
\end{figure*}

Figs.~\ref{fig:mnistresults0} through \ref{fig:celebaresults} include sample reconstruction results which are intended to qualitatively showcase the performance of the competing methods. The results in Figs.~\ref{fig:mnistresults0} and \ref{fig:mnistresults1} are arranged in groups of two rows of images, each group corresponding to reconstructions from the observed image (top row) and ground truth (bottom row) in the leftmost column of each image group. The remaining images in the top row of each group correspond to reconstructions by MisGAN, PBiGAN, MCFlow and PRFlow, respectively, from left to right. The bottom row in each group includes the mean squared error maps between the reconstruction by each method and the ground truth. Fig.~\ref{fig:mnistresults0} includes results across different rates of missing data.  It can be observed that, as the results from Table~\ref{tab:RMSE} indicate, GAN-based methods tend to produce higher MSE reconstructions. Further, the reconstructions produced by PRFlow showcase human-like handwriting across all levels, with strokes that are mostly continuous and largely uninterrupted.  Lastly, the images recovered by PRFlow almost always resemble a readable digit, which is not the case with the competing methods, particularly for missing rates of 80\% and above. 

Fig.~\ref{fig:mnistresults1} focuses on the 90\% missing data case and provides four additional examples.  As before, all of the images restored by PRFlow resemble human-like handwritten digits.  Failure to recover the original semantic content of the images happens mostly in cases where the original images themselves are ambiguous. Figs.~\ref{fig:cifarresults} and~\ref{fig:celebaresults} include reconstruction results on CIFAR-10 and CelebA. From left to right, the images include: ground truth, observed, and reconstructions by MNIST, PBiGAN, MCFlow and PRFlow. It can be seen that the Flow-based methods outperform the GAN-based methods, with PBiGAN lagging significantly behind. PRFlow has the overall edge in image quality with sharper edges, smoother backgrounds and more realistic reconstructions. Specifically, the edges of the plane and mountains against the sky are sharp in PRFlow reconstructions; the edges of sunglasses against skin are better defined; skin textures are more realistic and facial features (e.g., mouth, nose, hair strands) are rendered more naturally.  While there are similarities between the MCFlow and PRFlow renditions, there are edge sharpness and texture differences (e.g., ringing and blockiness artifacts being more pronounced in the MCFlow images) that likely lead to the measurable gap in performance showcased in Tables \ref{tab:RMSE}-\ref{tab:SCC}. Lastly, the bottommost row in Fig.~\ref{fig:celebaresults} illustrates a subtle but semantically significant reconstruction artifact where competing methods hallucinate a person with open eyes, while PRFlow accurately reconstructs a squinting face. We invite readers to attempt to fill in missing values themselves from the partially observed versions of the images. It can be a challenging task, in particular for high rates of missing data. We should note that humans have an advantage in that they know from experience what a number, an animal, or a face look like, whereas the algorithms competing herein were never exposed to a single fully observed image, and thus have to infer what the different objects look like by piecing together fractional observations from multiple images in the complete absence of labels. %Since no publicly available version of any of the methods intended for use with CIFAR was available, we extended the MNIST implementations to support CIFAR imagery. While MisGAN and MCFlow are well-behaved, PBiGAN's performance is lackluster, which likely means additional fine-tuning is needed. From our standpoint, the issues are not related to lack of convergence as training ran for 1000 epochs, and intermediate results showed that little additional learning took place beyond epoch 600. In contrast, PRFlow was found to converge much faster, usually within 100 epochs. 

\vspace{-0.2cm}
\section{Discussion}
\vspace{-0.2cm}

Traditionally, learning from incomplete or partially observed data has meant that trade-offs between various image quality aspects had to be incurred.  Specifically, prior methods on image imputation suffered at one or more of the following image quality attributes: (i) realism, (ii) pixel-level quality, and (iii) semantic consistency between the recovered and the partially observed image. These trade-offs became more significant as the degree of data paucity grew and approached unity. We hypothesize that this undesirable trend was due to the increasing level of ill-posedness of the recovery process and proposed a regularization approach that proved effective at addressing the three-pronged image quality trade-off. Extensive experimental results demonstrate that the proposed algorithm consistently matches or outperforms the performance of competing state-of-the-art approaches across all quality metrics in question. The seamless incorporation of domain knowledge in the form of a prior regularizer was made possible by the formulation of the learning task as a joint optimization objective.

{\small
\bibliographystyle{ieee_fullname}
\bibliography{egbib}

\begin{thebibliography}{10}\itemsep=-1pt

\bibitem{pmlr-v80-achlioptas18a}
Panos Achlioptas, Olga Diamanti, Ioannis Mitliagkas, and Leonidas Guibas.
\newblock Learning representations and generative models for 3{D} point clouds.
\newblock In Jennifer Dy and Andreas Krause, editors, {\em Proceedings of the
  35th International Conference on Machine Learning}, volume~80 of {\em
  Proceedings of Machine Learning Research}, pages 40--49, Stockholmsmässan,
  Stockholm Sweden, 10--15 Jul 2018. PMLR.

\bibitem{brock2018large}
Andrew Brock, Jeff Donahue, and Karen Simonyan.
\newblock Large scale {GAN} training for high fidelity natural image synthesis.
\newblock In {\em International Conference on Learning Representations}, 2019.

\bibitem{DBLP:conf/iclr/CohenS17}
Nadav Cohen and Amnon Shashua.
\newblock Inductive bias of deep convolutional networks through pooling
  geometry.
\newblock In {\em 5th International Conference on Learning Representations,
  {ICLR} 2017, Toulon, France, April 24-26, 2017, Conference Track
  Proceedings}. OpenReview.net, 2017.

\bibitem{doi:10.1111/1467-9574.00114}
A.~K. Dey and F.~H. Ruymgaart.
\newblock Direct density estimation as an ill-posed inverse estimation problem.
\newblock {\em Statistica Neerlandica}, 53(3):309--326, 1999.

\bibitem{Dinh2014NICENI}
Laurent Dinh, David Krueger, and Yoshua Bengio.
\newblock Nice: Non-linear independent components estimation.
\newblock {\em CoRR}, abs/1410.8516, 2014.

\bibitem{45819}
Laurent Dinh, Jascha Sohl{-}Dickstein, and Samy Bengio.
\newblock Density estimation using real {NVP}.
\newblock In {\em 5th International Conference on Learning Representations,
  {ICLR} 2017, Toulon, France, April 24-26, 2017, Conference Track
  Proceedings}. OpenReview.net, 2017.

\bibitem{DBLP:conf/iclr/DonahueKD17}
Jeff Donahue, Philipp Kr{\"{a}}henb{\"{u}}hl, and Trevor Darrell.
\newblock Adversarial feature learning.
\newblock In {\em 5th International Conference on Learning Representations,
  {ICLR} 2017, Toulon, France, April 24-26, 2017, Conference Track
  Proceedings}. OpenReview.net, 2017.

\bibitem{dumoulin2017adversarially-iclr}
Vincent Dumoulin, Mohamed Ishmael~Diwan Belghazi, Ben Poole, Alex Lamb, Martin
  Arjovsky, Olivier Mastropietro, and Aaron Courville.
\newblock Adversarially learned inference.
\newblock 2017.

\bibitem{engl1996regularization}
H.W. Engl, M. Hanke, and A. Neubauer.
\newblock {\em Regularization of Inverse Problems}.
\newblock Mathematics and Its Applications. Springer Netherlands, 1996.

\bibitem{Evgeniou2000}
T. Evgeniou, M. Pontil, and T. Poggio.
\newblock Regularization networks and support vector machines.
\newblock {\em Advances in Computational Mathematics}, 13(1), 2000.

\bibitem{10.5555/1756006.1859918}
Kuzman Ganchev, Jo\~{a}o Gra\c{c}a, Jennifer Gillenwater, and Ben Taskar.
\newblock Posterior regularization for structured latent variable models.
\newblock {\em J. Mach. Learn. Res.}, 11:2001–2049, Aug. 2010.

\bibitem{Georgeeaag2612}
Dileep George, Wolfgang Lehrach, Ken Kansky, Miguel L{\'a}zaro-Gredilla,
  Christopher Laan, Bhaskara Marthi, Xinghua Lou, Zhaoshi Meng, Yi Liu, Huayan
  Wang, Alex Lavin, and D.~Scott Phoenix.
\newblock A generative vision model that trains with high data efficiency and
  breaks text-based captchas.
\newblock {\em Science}, 358(6368), 2017.

\bibitem{NIPS2014_5423}
Ian Goodfellow, Jean Pouget-Abadie, Mehdi Mirza, Bing Xu, David Warde-Farley,
  Sherjil Ozair, Aaron Courville, and Yoshua Bengio.
\newblock Generative adversarial nets.
\newblock In Z. Ghahramani, M. Welling, C. Cortes, N.~D. Lawrence, and K.~Q.
  Weinberger, editors, {\em Advances in Neural Information Processing Systems
  27}, pages 2672--2680. Curran Associates, Inc., 2014.

\bibitem{DBLP:journals/corr/Goodfellow17}
Ian~J. Goodfellow.
\newblock {NIPS} 2016 tutorial: Generative adversarial networks.
\newblock {\em CoRR}, abs/1701.00160, 2017.

\bibitem{He2016DeepRL}
Kaiming He, X. Zhang, Shaoqing Ren, and Jian Sun.
\newblock Deep residual learning for image recognition.
\newblock {\em 2016 IEEE Conference on Computer Vision and Pattern Recognition
  (CVPR)}, pages 770--778, 2016.

\bibitem{heusel2017gans}
Martin Heusel, Hubert Ramsauer, Thomas Unterthiner, Bernhard Nessler, and Sepp
  Hochreiter.
\newblock {GAN}s trained by a two time-scale update rule converge to a local
  nash equilibrium.
\newblock In {\em Advances in neural information processing systems}, pages
  6626--6637, 2017.

\bibitem{hu-etal-2016-harnessing}
Zhiting Hu, Xuezhe Ma, Zhengzhong Liu, Eduard Hovy, and Eric Xing.
\newblock Harnessing deep neural networks with logic rules.
\newblock In {\em Proceedings of the 54th Annual Meeting of the Association for
  Computational Linguistics (Volume 1: Long Papers)}, pages 2410--2420, Berlin,
  Germany, Aug. 2016. Association for Computational Linguistics.

\bibitem{NIPS2018_8250}
Zhiting Hu, Zichao Yang, Russ~R Salakhutdinov, Lianhui Qin, Xiaodan Liang,
  Haoye Dong, and Eric~P Xing.
\newblock Deep generative models with learnable knowledge constraints.
\newblock In S. Bengio, H. Wallach, H. Larochelle, K. Grauman, N. Cesa-Bianchi,
  and R. Garnett, editors, {\em Advances in Neural Information Processing
  Systems 31}, pages 10501--10512. Curran Associates, Inc., 2018.

\bibitem{pmlr-v37-ioffe15}
Sergey Ioffe and Christian Szegedy.
\newblock Batch normalization: Accelerating deep network training by reducing
  internal covariate shift.
\newblock volume~37 of {\em Proceedings of Machine Learning Research}, pages
  448--456, Lille, France, 07--09 Jul 2015. PMLR.

\bibitem{4587659}
{Jian Sun}, {Zongben Xu}, and {Heung-Yeung Shum}.
\newblock Image super-resolution using gradient profile prior.
\newblock In {\em 2008 IEEE Conference on Computer Vision and Pattern
  Recognition}, pages 1--8, 2008.

\bibitem{NIPS2018_8224}
Durk~P Kingma and Prafulla Dhariwal.
\newblock Glow: Generative flow with invertible 1x1 convolutions.
\newblock In S. Bengio, H. Wallach, H. Larochelle, K. Grauman, N. Cesa-Bianchi,
  and R. Garnett, editors, {\em Advances in Neural Information Processing
  Systems 31}, pages 10215--10224. Curran Associates, Inc., 2018.

\bibitem{kingma2013autoencoding}
Diederik~P Kingma and Max Welling.
\newblock Auto-encoding variational bayes, 2013.
\newblock cite arxiv:1312.6114.

\bibitem{Kiran_2018}
B. Kiran, Dilip Thomas, and Ranjith Parakkal.
\newblock An overview of deep learning based methods for unsupervised and
  semi-supervised anomaly detection in videos.
\newblock {\em Journal of Imaging}, 4(2):36, Feb 2018.

\bibitem{NIPS2009_3707}
Dilip Krishnan and Rob Fergus.
\newblock Fast image deconvolution using hyper-laplacian priors.
\newblock In Y. Bengio, D. Schuurmans, J.~D. Lafferty, C.~K.~I. Williams, and
  A. Culotta, editors, {\em Advances in Neural Information Processing Systems
  22}, pages 1033--1041. Curran Associates, Inc., 2009.

\bibitem{krizhevsky2009learning}
Alex Krizhevsky, Geoffrey Hinton, et~al.
\newblock Learning multiple layers of features from tiny images.
\newblock Technical report, Citeseer, 2009.

\bibitem{Lecun98gradient-basedlearning}
Yann Lecun, Léon Bottou, Yoshua Bengio, and Patrick Haffner.
\newblock Gradient-based learning applied to document recognition.
\newblock In {\em Proceedings of the IEEE}, pages 2278--2324, 1998.

\bibitem{lecun-mnisthandwrittendigit-2010}
Yann LeCun and Corinna Cortes.
\newblock {MNIST} handwritten digit database.
\newblock 2010.

\bibitem{8099502}
C. {Ledig}, L. {Theis}, F. {Huszár}, J. {Caballero}, A. {Cunningham}, A.
  {Acosta}, A. {Aitken}, A. {Tejani}, J. {Totz}, Z. {Wang}, and W. {Shi}.
\newblock Photo-realistic single image super-resolution using a generative
  adversarial network.
\newblock In {\em 2017 IEEE Conference on Computer Vision and Pattern
  Recognition (CVPR)}, pages 105--114, 2017.

\bibitem{li2018learning}
Steven Cheng-Xian Li, Bo Jiang, and Benjamin Marlin.
\newblock {MisGAN:} learning from incomplete data with generative adversarial
  networks.
\newblock In {\em International Conference on Learning Representations}, 2019.

\bibitem{icml2020_3129}
Steven Cheng-Xian Li and Benjamin Marlin.
\newblock Learning from irregularly-sampled time series: A missing data
  perspective.
\newblock In {\em Proceedings of Machine Learning and Systems 2020}, pages
  5756--5765. 2020.

\bibitem{liu2015faceattributes}
Ziwei Liu, Ping Luo, Xiaogang Wang, and Xiaoou Tang.
\newblock Deep learning face attributes in the wild.
\newblock In {\em Proceedings of International Conference on Computer Vision
  (ICCV)}, December 2015.

\bibitem{Ma2018PartialVF}
C. Ma, Wenbo Gong, Jos{\'e}~Miguel Hern{\'a}ndez-Lobato, Noam Koenigstein,
  Sebastian Nowozin, and C. Zhang.
\newblock Partial vae for hybrid recommender system.
\newblock 2018.

\bibitem{DBLP:journals/corr/abs-1801-00631}
Gary Marcus.
\newblock Deep learning: {A} critical appraisal.
\newblock {\em CoRR}, abs/1801.00631, 2018.

\bibitem{pmlr-v97-mattei19a}
Pierre-Alexandre Mattei and Jes Frellsen.
\newblock {MIWAE}: Deep generative modelling and imputation of incomplete data
  sets.
\newblock In Kamalika Chaudhuri and Ruslan Salakhutdinov, editors, {\em
  Proceedings of the 36th International Conference on Machine Learning},
  volume~97 of {\em Proceedings of Machine Learning Research}, pages
  4413--4423, Long Beach, California, USA, 09--15 Jun 2019. PMLR.

\bibitem{miyato2018spectral}
Takeru Miyato, Toshiki Kataoka, Masanori Koyama, and Yuichi Yoshida.
\newblock Spectral normalization for generative adversarial networks.
\newblock In {\em International Conference on Learning Representations}, 2018.

\bibitem{8621955}
N. {Muralidhar}, M.~R. {Islam}, M. {Marwah}, A. {Karpatne}, and N.
  {Ramakrishnan}.
\newblock Incorporating prior domain knowledge into deep neural networks.
\newblock In {\em 2018 IEEE International Conference on Big Data (Big Data)},
  pages 36--45, 2018.

\bibitem{10.1162/neco.1992.4.4.473}
Steven~J. Nowlan and Geoffrey~E. Hinton.
\newblock Simplifying neural networks by soft weight-sharing.
\newblock {\em Neural Comput.}, 4(4):473–493, July 1992.

\bibitem{oord2016wavenet}
Aaron van~den Oord, Sander Dieleman, Heiga Zen, Karen Simonyan, Oriol Vinyals,
  Alex Graves, Nal Kalchbrenner, Andrew Senior, and Koray Kavukcuoglu.
\newblock Wavenet: A generative model for raw audio, 2016.
\newblock cite arxiv:1609.03499.

\bibitem{NEURIPS2019_5f8e2fa1}
Ali Razavi, Aaron van~den Oord, and Oriol Vinyals.
\newblock Generating diverse high-fidelity images with vq-vae-2.
\newblock In H. Wallach, H. Larochelle, A. Beygelzimer, F. d\textquotesingle
  Alch\'{e}-Buc, E. Fox, and R. Garnett, editors, {\em Advances in Neural
  Information Processing Systems}, volume~32. Curran Associates, Inc., 2019.

\bibitem{pmlr-v32-rezende14}
Danilo~Jimenez Rezende, Shakir Mohamed, and Daan Wierstra.
\newblock Stochastic backpropagation and approximate inference in deep
  generative models.
\newblock In Eric~P. Xing and Tony Jebara, editors, {\em Proceedings of the
  31st International Conference on Machine Learning}, volume~32 of {\em
  Proceedings of Machine Learning Research}, pages 1278--1286, Bejing, China,
  22--24 Jun 2014. PMLR.

\bibitem{Richardson_2020_CVPR}
Trevor~W. Richardson, Wencheng Wu, Lei Lin, Beilei Xu, and Edgar~A. Bernal.
\newblock {MCF}low: Monte carlo flow models for data imputation.
\newblock In {\em Proceedings of the IEEE/CVF Conference on Computer Vision and
  Pattern Recognition (CVPR)}, June 2020.

\bibitem{NIPS2004_2722}
Lorenzo Rosasco, Andrea Caponnetto, Ernesto~D. Vito, Francesca Odone, and
  Umberto~D. Giovannini.
\newblock Learning, regularization and ill-posed inverse problems.
\newblock In L.~K. Saul, Y. Weiss, and L. Bottou, editors, {\em Advances in
  Neural Information Processing Systems 17}, pages 1145--1152. MIT Press, 2005.

\bibitem{rosenblatt1956}
Murray Rosenblatt.
\newblock Remarks on some nonparametric estimates of a density function.
\newblock {\em Ann. Math. Statist.}, 27(3):832--837, 09 1956.

\bibitem{conf/cvpr/SchroffKP15}
Florian Schroff, Dmitry Kalenichenko, and James Philbin.
\newblock Facenet: A unified embedding for face recognition and clustering.
\newblock In {\em CVPR}, pages 815--823. IEEE Computer Society, 2015.

\bibitem{10.5555/2627435.2670313}
Nitish Srivastava, Geoffrey Hinton, Alex Krizhevsky, Ilya Sutskever, and Ruslan
  Salakhutdinov.
\newblock Dropout: A simple way to prevent neural networks from overfitting.
\newblock {\em J. Mach. Learn. Res.}, 15(1):1929–1958, Jan. 2014.

\bibitem{5539933}
Y. {Tai}, S. {Liu}, M.~S. {Brown}, and S. {Lin}.
\newblock Super resolution using edge prior and single image detail synthesis.
\newblock In {\em 2010 IEEE Computer Society Conference on Computer Vision and
  Pattern Recognition}, pages 2400--2407, 2010.

\bibitem{TakahashiMCMC}
M. Takahashi.
\newblock Statistical inference in missing data by mcmc and non-mcmc multiple
  imputation algorithms: Assessing the effects of between-imputation
  iterations.
\newblock {\em Data Science Journal}, 16(37):1--17, 2017.

\bibitem{10.1145/3186549.3186562}
Niket Tandon, Aparna~S. Varde, and Gerard de Melo.
\newblock Commonsense knowledge in machine intelligence.
\newblock {\em SIGMOD Rec.}, 46(4):49–52, Feb. 2018.

\bibitem{Tikhonov/Arsenin/77}
A.~N. Tikhonov and V.~Y. Arsenin.
\newblock {\em Solutions of Ill-posed problems}.
\newblock W.H.~Winston, 1977.

\bibitem{Ulyanov_2018_CVPR}
Dmitry Ulyanov, Andrea Vedaldi, and Victor Lempitsky.
\newblock Deep image prior.
\newblock In {\em Proceedings of the IEEE Conference on Computer Vision and
  Pattern Recognition (CVPR)}, June 2018.

\bibitem{Vapnik1998}
Vladimir~N. Vapnik.
\newblock {\em Statistical Learning Theory}.
\newblock Wiley-Interscience, 1998.

\bibitem{7410407}
Z. {Wang}, D. {Liu}, J. {Yang}, W. {Han}, and T. {Huang}.
\newblock Deep networks for image super-resolution with sparse prior.
\newblock In {\em 2015 IEEE International Conference on Computer Vision
  (ICCV)}, pages 370--378, 2015.

\bibitem{DBLP:journals/corr/abs-1809-04747}
Tao Yang, Georgios Arvanitidis, Dongmei Fu, Xiaogang Li, and S{\o}ren Hauberg.
\newblock Geodesic clustering in deep generative models.
\newblock {\em CoRR}, abs/1809.04747, 2018.

\bibitem{yang2017deep}
Xitong Yang, Palghat Ramesh, Radha Chitta, Sriganesh Madhvanath, Edgar~A
  Bernal, and Jiebo Luo.
\newblock Deep multimodal representation learning from temporal data.
\newblock In {\em Proceedings of the IEEE Conference on Computer Vision and
  Pattern Recognition}, pages 5447--5455, 2017.

\bibitem{pmlr-v80-yoon18a}
Jinsung Yoon, James Jordon, and Mihaela van~der Schaar.
\newblock {GAIN}: Missing data imputation using generative adversarial nets.
\newblock In Jennifer Dy and Andreas Krause, editors, {\em Proceedings of the
  35th International Conference on Machine Learning}, volume~80 of {\em
  Proceedings of Machine Learning Research}, pages 5689--5698,
  Stockholmsmässan, Stockholm Sweden, 10--15 Jul 2018. PMLR.

\bibitem{zhang2020super}
Yuxin Zhang, Zuquan Zheng, and Roland Hu.
\newblock Super resolution using segmentation-prior self-attention generative
  adversarial network, 2020.

\end{thebibliography}
}

\end{document}